\PassOptionsToPackage{hyphens}{url}

\documentclass[11pt,a4paper,hyphens]{article}
\usepackage{times,latexsym}

\usepackage{hyperref}
\usepackage[hyphens]{url}
\usepackage{breakurl}

\usepackage{authblk} 

\usepackage[acceptedWithA]{tacl2021v1}

\usepackage{xspace,mfirstuc,tabulary}

\newcommand{\enotesoff}{\long\gdef\enote##1{}}
\newcommand{\enoteson}{\long\gdef\enote##1{\par\noindent\fbox{\parbox {\textwidth}{{\large DRAFT.} \small\scshape ##1}}\\[0.3ex]}}
\enotesoff
\enoteson

\newcommand{\geoparnoteson}{\long\gdef\geoparnote##1{\par\noindent\fbox{\parbox {.9\columnwidth}{{\color{blue} geopar comment:} \small\scshape ##1}}\\[0.1ex]}}
\geoparnoteson

\newif\iftaclinstructions
\taclinstructionsfalse 
\iftaclinstructions
\renewcommand{\confidential}{}
\renewcommand{\anonsubtext}{GAsfs}
\newcommand{\instr}
\fi

\iftaclpubformat 

\else

\fi

\usepackage[T1]{fontenc}
\usepackage[greek,english]{babel}
\languageattribute{greek}{keep-semicolon}

        \hypersetup{
           breaklinks=true,   
           colorlinks=true,   
           pdfusetitle=true,  
        }

\title{Meltemi: The first open Large Language Model for Greek}

\author{
Leon Voukoutis, Dimitris Roussis, Georgios Paraskevopoulos, \\Sokratis Sofianopoulos, Prokopis Prokopidis,    Vassilis Papavasileiou, \\ Athanasios Katsamanis, Stelios Piperidis, Vassilis Katsouros
  \ \\
  \\
  Institute for Speech and Language Processing, Athena Research Center\\ Artemidos 6 \& Epidavrou, Athens, Greece\\
  \texttt{vsk@athenarc.gr}
}

\begin{document}

\maketitle

\begin{abstract}
  We describe the development and capabilities of Meltemi 7B, the first open Large Language Model for the Greek language. Meltemi 7B has 7 billion parameters and is trained on a 40 billion token Greek corpus. For the development of Meltemi 7B, we adapt Mistral, by continuous pretraining on the Greek Corpus.
  Meltemi 7B contains up-to-date information up to September 2023.
  Furthermore, we have translated and curated a Greek instruction corpus, which has been used for the instruction-tuning of a chat model, named Meltemi 7B Instruct. Special care has been given to the alignment and the removal of toxic content for the Meltemi 7B Instruct. The developed models are evaluated on a broad set of collected evaluation corpora, and examples of prompts and responses are presented. Both Meltemi 7B and Meltemi 7B Instruct are available\footnote{\url{https://huggingface.co/ilsp/}} under the Apache 2.0 license.
\end{abstract}

\section{Introduction}\label{sec:intro}

Large Language Models (LLMs) have emerged as potent tools, that enable user-facing Artificial Intelligence (AI) applications. 
Their success lies in their ability to understand and generate natural language, which provides a natural and intuitive avenue for users to interact with them. Nevertheless, their training requires the dedication of massive amounts of resources, both in terms of computing power, as well as data collection and curation. Thus, LLM development progress has favored major languages (e.g. English, Chinese), because their large user bases justify the significant upfront investment.

There is a growing need to overcome this barrier to entry because the commoditization of AI is rapidly driving its evolution from a novelty to a core utility\footnote{\url{https://www2.deloitte.com/us/en/pages/consulting/articles/the-future-of-ai.html}}; akin to the evolution of the internet during the late nineties. The impact of native, widely available, high-performant AI foundation models thus, is a mandate that should be incorporated into the strategic plan of local stakeholders.

Continual pretraining of LLMs has emerged as a significant area of research aimed at exploiting existing models by extending the model to languages other than English.
Ǎguila-7B \cite{Aguilla2023} is a 7B parameters model based on Falcon-7b \citep{Falcon2023} and further trained with a mixture of Spanish, Catalan and English data. Using a similar strategy, \citet{Pluster2023} released LeoLM-7B, a German foundation model created by applying continual pretraining on Llama2-7B. Finally, similar work has been performed for Japanese with \citet{rakutenai7b} and \citet{tokyotechllm2024}, both being foundation models created by applying continual pretraining on Mistral-7B.

In this paper, we present the development of Meltemi 7B and Meltemi 7B Instruct, the first dedicated open LLM for the Greek language, and the corresponding instruction following model. 
Furthermore, we develop a comprehensive benchmark for LLM evaluation in the Greek language. This benchmark covers a wide range of tasks, i.e., translation, dialogue, reasoning, and consists of translated and curated versions of existing test sets. The development of LLMs  brings to light a necessity for a plan to be formulated for their maintenance and continuous provision. We initiate this discussion for the environmentally and economically sustainable deployment, continuous data integration, and updating of the developed models.












\section{Our Method}\label{sec:materials}


LLMs require significant amounts of data in order to learn how to generate fluent texts as well as capture culture-specific knowledge. Therefore, adapting an LLM for the Greek language requires addressing its underrepresentation in datasets comprising trillions of tokens for other languages, such as RedPajama\footnote{\url{https://github.com/togethercomputer/RedPajama-Data}}. Notable approaches to tackle this challenge include the creation of synthetic data for a low-resource language via the translation of English corpora, as for example in Jais, a bilingual model pretrained for Arabic and English from scratch \cite{sengupta2023jais}. There have also been several approaches to extend the pretraining of English-centric LLMs, such as the Sabi{\'a} models for Portuguese \cite{10.1007/978-3-031-45392-2_15}, BgGPT for Bulgarian \cite{INSAIT2024}. 

Existing resources for the Greek language which have been processed for direct use in LLMs include CulturaX \cite{nguyen2023culturax} and the HPLT corpus \cite{de2024new}.

In the subsections that follow, we detail out strategy for creating and evaluating Meltemi:

\begin{itemize}
  \item Acquiring, collecting, filtering, preprocessing, and deduplicating large amounts of Greek texts from various diverse sources.
  \item Extending the original tokenizer with Greek tokens and fine-tuning the newly initialized embeddings on a smaller subcorpus.
  \item Continually pretraining Mistral 7B \cite{jiang2023mistral} with a mix of Greek, English, and Parallel (English-Greek) data.
  \item Imbuing the new foundation model with chat capabilities using the ORPO \cite{hong2024reference} algorithm and translated preference data.
  \item Evaluating Meltemi and Meltemi 7B Instruct on Greek benchmarks, including novel, existing, and synthetically generated test sets.
\end{itemize}

\subsection{Pretraining Data}\label{sec:corpus}

As mentioned earlier, LLMs require significant amounts of pretraining data in order to form a strong foundation model which can be then finetuned and aligned to human preferences. More data is not always better; data quality is also of utmost importance. In our work, we set the goal of sourcing as much Greek monolingual texts as possible, as well as to ensure that the pretraining data mix is of high quality, by using various filtering and pre-processing procedures.

Our continual pretraining approach adapts an existing model for a new natural language using data which induce a strong distribution shift to the model, as it has already been trained with dissimilar data\footnote{Note that there is no published information regarding the data that Mistral 7B has been trained on, but it is evident that it has been predominantly trained on English texts.}. Taking into consideration this large distribution shift which could potentially lead to catastrophic forgetting, our approach involves re-warming and re-decaying the learning rate \cite{ibrahim2024simple}, and utilizing both Greek and English monolingual data in the continual pretraining data mix.

Regarding the Greek monolingual data, we leveraged original Greek monolingual texts from various sources, including Wikipedia, ELRC-SHARE \cite{losch2021collection}, EUR-LEX \& MultiEUR-LEX \cite{chalkidis-etal-2019-large, chalkidis-etal-2021-multieurlex}, MaCoCu \cite{non-etal-2022-macocu}, CLARIN-EL \citep{GavriilidouEtAl2023}, parliamentary proceedings \cite{erjavec2023parlamint}, full texts (i.e., theses, dissertations, etc.) and abstracts from Greek academic repositories, as well as pre-filtered resources originally compiled from the web, such as CulturaX \cite{nguyen2023culturax}.

In order to create an English monolingual corpus, we decided to use the English counterparts of Wikipedia and EUR-LEX, as well as English full texts and abstracts from academic records in Greek repositories. Additionally, we included the AutoMathText dataset \cite{zhang2024automathtext}, which is a collection of math-related documents originating from web data, papers on arXiv, and code/notebooks on GitHub. These documents have undergone an automatic selection process using Qwen-72B \cite{bai2023qwen} which resulted in a score between 0 and 1, reflecting the relevancy to the mathematical domain and the educational value of each document. In our work, we select all documents with a score greater than or equal to 0.70.

Furthermore, we also include augmented English-Greek translation data in the pretraining corpus, randomly sampled for each translation direction, i.e., EN-EL and EL-EN. Adding translation data to the continual pretraining corpus improves translation quality \cite{alves2024tower}, while there also has been limited empirical evidence which suggests that LLMs address multilingualism by first translating queries into English, process them using English and their multilingual knowledge, and then translate the responses back into the original language \cite{zhao2024large}. The parallel data corpus is a diverse mixture of selected high-quality datasets which we have acquired in previous work \cite{losch2021collection,roussis2022scipar,roussis2022constructing}, or are available in ELRC-SHARE \& OPUS \cite{tiedemann2012parallel}.

In order to preprocess and clean the monolingual texts, we first had to ensure that they shared a common format and, thus, we extracted textual content from documents that were stored in PDF and HTML formats, we converted each textual document (or datapoint) to a JSON object which includes the document's textual content, as well as other metadata such as the text's language, its number of words, its dataset of origin, any source URLs (if available), etc.

After extensive manual inspection of each source dataset, we decided to use different pre-processing and filtering pipelines, as we observed large variations in data quality, readiness, and cleanliness, which were also connected with the way the dataset was originally acquired. For example, we removed documents from an in-house list of -approximately 500- Greek blacklisted URLs, if metadata about the source URL were available, as in CulturaX. Also, in documents extracted from PDFs, we used regex patterns to remove textual parts with glued words or many consecutive single alphabetic characters, which are usually artifacts of the PDF extraction process.

Other rule-based filters that were used include:
\begin{itemize}
  \item Removal of small documents (<300 characters or <6 words)
  \item Removal of documents with extremely long words (at least one with >60 characters)
  \item Removal of documents with at least 2 bad words from a curated in-house list with Greek bad words and phrases.
  \item Removal of documents containing the "lorem ipsum" substring.
  \item Removal of noisy parallel sentences in the EN-EL parallel dataset using an in-house filtering pipeline \cite{papavassiliou2018ilsp,roussis2022arc}.
\end{itemize}

Our filtering pipeline also made use of more sophisticated tools, such as Monocleaner \footnote{\url{https://github.com/bitextor/monocleaner}}, which assigns a fluency score in a text's paragraphs using a 7-gram KenLM language model \cite{prompsit:2018:WMT}. A threshold of 0.7 was set for the score given by Monocleaner for documents such as the ones extracted from PDFs of academic records. These scores were already available for MaCoCu, while 5-gram KenLM language models trained on Wikipedia were used in CulturaX \cite{nguyen2023culturax}. Likewise, parallel corpora were filtered using a LASER \cite{artetxe2018margin,artetxe2019massively} threshold of 1.06 where available, as well as with a threshold of 0.7 of the score calculated using BiCleaner AI \cite{zaragoza-bernabeu-etal-2022-bicleaner} full model for English-Greek, which is a tool for detecting noisy sentence pairs in parallel corpora.

Finally, we deduplicated the pretraining corpus by utlizing the MinHashLSH near-deduplication method \cite{broder1997resemblance,leskovec2020mining} which has been commonly used for deduplicating large datasets at scale \cite{penedo2023refinedweb, nguyen2023culturax, tokpanov2024zyda, penedo2024fineweb}. In particular, we followed a 2-stage process:
\begin{itemize}
  \item \textbf{Intra-dataset dedulication:}
  First, we used MinHashLSH to identify duplicates within each dataset which was not known to be already deduplicated.
  \item \textbf{Cross-dataset dedulication:}
  Then, we removed duplicates across the whole corpus which emerged after concatenating all of the datasets with the application of MinHashLSH.
\end{itemize}

In particular, we adapted the implementation in the text-dedup\footnote{\url{https://github.com/ChenghaoMou/text-dedup/}} repository for our case by using 5-gram subsets, a MinHash signature of 128, and a Jaccard similarity threshold of 0.8; a choice of parameters which has been based on the approach used for deduplicating CulturaX \cite{nguyen2023culturax}, although other choices have also been shown to work well.

In order to identify duplicates in the parallel corpora, we followed a different approach. Initially, we converted all sentence pairs to lowercase and removed digits and punctuation. Afterwards, we removed duplicate sentence pairs based on either the source or the target sentence, by ensuring no Greek or English sentence appears more than once in the combined parallel dataset \cite{roussis2022arc}.

The resulting high-quality continual pretraining corpus consists of approximately 54.5 billion tokens as listed in Table \ref{table:pretraining-data}.

\begin{table}[t]
  \centering

\begin{tabular}{c|r|c}
\hline
Subcorpus & No. of Tokens & Percentage \\
\hline
Greek & 43,383,244,502 & 79.5\% \\
English & 10,538,413,259 & 19.3\% \\
Parallel & 633,816,023 & 1.2\% \\
\hline
\textbf{Total} & \textbf{54,555,473,784} & \textbf{100\%} \\
\hline
\end{tabular}

\caption{Continual Pretraining Corpus.}
\label{table:pretraining-data}
\end{table}

\subsection{Tokenizer and Embeddings Expansion}\label{sec:tokenizer}

In this section, we present our experiments and methodology towards extending the original tokenizer of Mistral 7B to facilitate efficient handling of Greek textual data.

The original tokenizer of Mistral 7B v0.1 \cite{jiang2023mistral} is comprised of 32,000 tokens and is generally inefficient for Greek texts. To ascertain this, we conducted preliminary tests on diverse Greek and English corpora (approximately 2M words each) and calculated the average amount of tokens produced by the tokenizer for each word, also known as fertility. The fertility of a tokenizer is an extremely important quantity, as higher values directly correlate with higher training and inference costs; a phenomenon observed for various underrepresented natural languages \cite{csaki2023efficiently}.

\begin{table}[t]
  \centering

\begin{tabular}{p{0.3\linewidth}|p{0.15\linewidth}|p{0.15\linewidth}|p{0.15\linewidth}}
\hline
Tokenizer Model & Vocab. Size & Fertility Greek & Fertility English \\
\hline
Mistral 7B & 32,000 & 6.80 & 1.49 \\
Meltemi 7B & 61,362 & \textbf{1.52} & \textbf{1.44} \\
\hline
\end{tabular}

\caption{Tokenizer Statistics.}
\label{table:tokenizers}
\end{table}

As we can observe in Table \ref{table:tokenizers}, the original tokenizer of Mistral 7B produces approximately 4.4 more tokens per word for Greek texts. This would mean that, on average, 4.4 times more computing resources would be needed for training and inference. Therefore, increasing the vocabulary size of the original tokenizer proved to be a vital step in significantly reducing training costs and increasing inference speed.

Following the expansion of the initial Mistral-7B tokenizer to 61,362 tokens we expanded the embedding layer of the base model to match that number and then rounded the size up to a multiple of 8 for computational efficiency. 

\subsection{Foundation Model Training}\label{sec:training}

Our pretraining procedure is split into two stages. We first loosely train only the freshly initialized embeddings, keeping therest of the parameters frozen and subsequently unfreeze the rest of the model, warm restart it and continue with training all the parameters. For both stages of this procedure we utilized 8x NVIDIA H100 GPUs and Deepspeed Zero \citep{DeepSpeed2020} for distributed training.

In stage 1 the embeddings for each new token are first initialized from the average of the embeddings of the tokens that would be retrieved by the initial tokenizer, when prompted to encode the given token. The same procedure was followed for the LM head respectively. Subsequently the added embeddings and LM head were trained on a small subset of the dataset with all other parameters frozen, to be brought in line with the rest of the weights. For this stage we trained for 2,500 steps with a maximum learning rate of $2.5\mathrm{e}{-4}$, a linear warmup over 250 steps, followed by a cosine decay to a minimum of $2.5\mathrm{e}{-5}$ and an effective batch size of 1.5M tokens. The optimizer used was the AdamW optimizer \citep{adamw2017} with $\beta_1 = 0.9, \beta_2 = 0.999, \epsilon = 10^{-5}$ and gradient clipping set to $1.0$. 

In stage 2 we warm restart the model, training over 24,800 steps with an effective batch size of 4.5M tokens. We set a maximum learning rate of $2.5\mathrm{e}{-5}$ that is reached after a linear warmup over 248 steps. We employ a cosine decay schedule of the learning rate down to a minimum of $2.5\mathrm{e}{-6}$, where it plateaus over the final 10\% of training steps. The optimizer used was the AdamW optimizer \citep{adamw2017} with $\beta_1 = 0.9, \beta_2 = 0.95, \epsilon = 10^{-5}$ and gradient clipping set to 1.0.


\begin{table*}[t]
  \centering

\begin{tabular}{p{0.2\linewidth} | c | p{0.65\linewidth}}
\hline
Name  & \# Ex. & Description \\
\hline
\href{https://huggingface.co/datasets/ilsp/arc_greek}{ARC Greek} & 7.78K & MT of ARC \cite{allenai:arc}, a dataset of science exam questions (with typically four answer options) partitioned into a Challenge and an Easy Set of 2.6K and 5.2 questions. \\
\href{https://huggingface.co/datasets/ilsp/truthful_qa_greek}{Truthful QA Greek} & 817 & Edited MT of Truthful QA \cite{lin-etal-2022-truthfulqa}, a dataset of questions that are crafted so that some humans would answer wrongly due to a false belief or misconception. \\
\href{https://huggingface.co/datasets/ilsp/hellaswag_greek}{HellaSwag Greek} & 59.8K & MT of the HellaSwag dataset \cite{zellers-etal-2019-hellaswag} for commonsense NLI. \\
\href{https://huggingface.co/datasets/ilsp/mmlu_greek}{MMLU Greek} & 15.9K & MT of the MMLU dataset \cite{hendrycks2021measuring} of multiple-choice questions from 57 tasks including elementary mathematics, history, computer science, law, etc. \\
\href{https://huggingface.co/datasets/facebook/belebele/viewer/default/ell_Grek}{Belebele (ell)} & 900 & The Greek part of Belebele \cite{bandarkar2023belebele}, a multiple-choice machine reading comprehension dataset covering 122 language variants. \\
\href{https://huggingface.co/datasets/ilsp/medical_mcqa_greek}{Greek Medical Multiple Choice QA} & 2.03K & Multiple choice questions extracted from past medical exams of the Greek National Acadenic Recognition and Information Center available at {\small \url{https://www.doatap.gr}}. \\
\hline
\end{tabular}

\caption{Greek evaluation datasets}
\label{table:eval_datasets}
\end{table*}

\begin{table*}[t]
  \centering

\begin{tabular}
{c|c|c|c}
\hline
\multicolumn{1}{c|}{}&\multicolumn{3}{c}{Models} \\
\hline
Task & Mistral-7B-v0.1 &  Meltemi-7B &  Meltemi-7B-Instruct \\
\hline
ARC-C Greek & 27.22 & \textbf{47.17} & 40.8 \\
TruthfulQA:MC2 Greek & 44.93 & 45.19 & \textbf{53.8} \\
HellaSwag Greek & 35.20 & \textbf{65.75} & 63.7 \\
MMLU Greek & 28.35 & 42.45 & \textbf{45.9} \\
Belebele (ell) & 35.77 & 68.66 & \textbf{75.5} \\
Greek Medical MC QA & 27.77 & \textbf{48.12} & 48.0 \\
\hline
Average Greek & 33.20 & 52.89 & \textbf{54.6} \\
\hline
ARC-C & \textbf{59.98} & 54.26 & 56.6 \\
TruthfulQA:MC2 & 42.15 & 40.60 & \textbf{51.2} \\
HellaSwag & \textbf{83.31} & 79.60 & 78.3 \\
MMLU & \textbf{64.16} & 56.86 & 57.0 \\
Winogrande & \textbf{78.37} & 73.16 & 70.2 \\
GSM8K (pass@1) & \textbf{34.50} & 22.13 & 32.8 \\
\hline
Average English & \textbf{60.41} & 54.43 & 57.7 \\
\hline
\end{tabular}

\caption{Evaluation Results.}
\label{table:eval_results}
\end{table*}

\subsection{Instruction Tuning}\label{sec:instruction_tuning}


In order to create the Meltemi 7B Instruct version of our model, we used ORPO \cite{hong2024reference} to align our model with human preferences. The implementation is based on the TRL library from Huggingface \citep{vonwerra2022trl} and partially on the Alignment Handbook repository \citep{alignment_handbook2023}. In particular, we translated a mix of 12 preference datasets from Huggingface and carefully curated them by:
\begin{itemize}
  \item Using preference data in which the chosen response has a high rating (no ties).
  \item Fixing formatting errors and filtering out cases with translation inconsistencies, excessive Unicode.
  \item Adding appropriate Greek system messages to each dataset.
\end{itemize}

Finally, we used the high quality translated Greek preference dataset comprising 89,730 preference triplets in Greek and 7,342 preference triplets in English from the initial datasets to mirror the initial training distribution, to create Meltemi 7B Instruct. Following common practice, we use special tokens to format the dataset for chat use, by denoting different roles: system messages, user messages, and assistant messages. We randomly apply custom Greek system messages to data without a pre-existing system role, while we also utilize RAG-specific, CoT-specific, mathematics-specific, and coding-specific system messages for the corresponding data. 

Training on the 97,072 preference data ran for 2 epochs over 2 days on 4x NVIDIA RTX A6000 GPUs utilizing Deepspeed Zero stage 3, with a maximum learning rate of $1\mathrm{e}{-6}$, that is reached after 110 steps of linear warmup followed by cosine decay to 0.

\subsection{Evaluation}\label{sec:evaluation-corpora}


Our evaluation suite includes Greek machine-translated versions of established English benchmarks for language understanding and reasoning, publicly available QA benchmarks targetting Greek (Belebele (ell) (8-shot), Greek Medical Multiple Choice QA (15-shot)) , and a novel benchmark with questions extracted from past medical exams as described in Table \ref{table:eval_datasets}. We also evaluate the effect of the continual pretraining on the English capabilities of the model on the \href{https://huggingface.co/spaces/open-llm-leaderboard/open_llm_leaderboard}{OpenLLMLeaderboard} tasks. We implemented our evaluation procedure based on a \href{https://github.com/LeonVouk/lighteval}{fork} of the \texttt{lighteval} framework \cite{lighteval}. We follow the same experimental setup as the OpenLLMLeaderboard for the English tasks and their machine-translated Greek counterparts. 

We report our evaluation results in Table \ref{table:eval_results}. We see that Meltemi-7B enhances performance across all Greek test sets by a +20.2\% average improvement. On the other side, the model performs worse than Mistral-7B for the English tasks, trailing by -6\%. The difference in style between Greek and English data the model was initially trained on and the capacity of a 7B model, has impacted the performance in English tasks. This issue is partially fixed after the models alignment with human preferences.

The same tendency has been observed for similar efforts in other languages. \citet{Pluster2023} report that LeoLM-7B achieved an average improvement of +4.8\% for German benchmarks compared to the base Llama-2 model, while it achieved lower average scores by -2.6\% on English benchmarks. Swallow-MS-7b-v0.1 \citep{tokyotechllm2024}, has shown an average improvement of +8\% on Japanese benchmarks versus its base model, and lower average scores of -5.3\% on English benchmarks.












\section{Discussion and Conclusions}\label{sec:conclusions}

In this paper we have extended Mistral-7B using continuous pretraining to create Meltemi-7B, the first open Large Language Model for the Greek language. Meltemi was further tuned on instruction data using state-of-the-art preference optimization techniques, yielding the chat model Meltemi 7B Instruct. For evaluation we have created a benchmark suite in Greek by translating and curating a set of popular English benchmarks and including a novel dataset targeted towards medical question answering. 
Our evaluation indicates that Meltemi 7B Instruct achieves state-of-the-art results, indicating a $20.2\%$ absolute performance increase for the Greek language averaged across metrics and datasets over Mistral 7B, with a moderate $6\%$ drop for the English benchmarks.

We hope that this effort furthers the discussion about the need for open-source LLMs for mid and low-resource languages, which take into account cultural aspects, local laws and customs. We acknowledge that this requires a significant economic and environmental investment. For reference, the development of Meltemi 7B required an energy consumption of $2300$ kWh. While continual pretraining helps to reduce the cost of extending existing models to new languages and domains, we want to further investigate techniques for efficient model adaptation. Furthermore, in the future we want to explore the adaptation of larger models for Greek, with a focus on sustainability, and also explore multimodality, thus creating  a suite of open and multimodal models for the Greek language.


\section*{Acknowledgements}\label{sec:acknowledgements}

The authors wish to thank AWS and GRNET, especially Nikiforos Botis and Panos Louridas, for their ongoing support and helping us attain the required training infrastructure. We also acknowledge the support of the Luleå University of Technology for providing computational resources, thus supporting part of our experiments. Many thanks to Olga Yannoutsou for her help in the translation of one of the evaluation datasets. Special thanks to all members of the Institute for Language and Speech Processing, Athena RC who participated in the internal evaluations and especially to Dimitris Galanis, Maria Giagkou, Sotiris Kotitsas and Dimitris Pappas for their valuable advice. Finally we are grateful to our colleagues in Xanthi who provided us access to four NVIDIA RTX A6000 GPUs for internal validation.

\bibliographystyle{acl_natbib}
\bibliography{tacl2021}

\appendix

\section{Examples of queries and responses}
\selectlanguage{greek}
\begin{description}
    \item[\textlatin{User}:] Πώς πιστεύεις θα είναι η ζωή σε 100 χρόνια;
    \item[\textlatin{Meltemi}:] Ως μοντέλο γλώσσας AI, δεν μπορώ να προβλέψω το μέλλον με βεβαιότητα ή προκατάληψη. Ωστόσο, εδώ είναι μερικές πιθανές προβλέψεις που βασίζονται στο τρέχον επίπεδο τεχνολογικής προόδου και στην κοινωνική ανάπτυξη: 
    
    Η τεχνητή νοημοσύνη μπορεί να γίνει πιο προηγμένη και ενσωματωμένη στην καθημερινή μας ζωή. Οι εικονικοί βοηθοί και οι \textlatin{chatbot} μπορούν ενδεχομένως να αντικαταστήσουν πολλές εργασίες ρουτίνας και ακόμη και ορισμένες θέσεις εργασίας. Μπορεί επίσης να δούμε την εμφάνιση αυτόνομων μηχανών που μπορούν να εκτελούν πολύπλοκες λειτουργίες σε βιομηχανίες όπως η κατασκευή, η γεωργία και οι μεταφορές\ldots

    \item[\textlatin{User}:] Ποιος ήταν ο πρώτος που πραγματοποίησε πρακτικά, το 1803, την πρόωση των πλοίων με ατμοκίνηση;
    \item[\textlatin{Meltemi:}] Ο πρώτος που πραγματοποίησε πρακτικά την προώση των πλοίων με ατμοκίνηση ήταν ο Ρόμπερτ Φούλτον.

    \item[\textlatin{User:}] Τα εμβόλια-υπομονάδες περιέχουν\\α. πρωτεΐνες με αντιγονική δράση.\\β. γυμνό \textlatin{DNA} του μικροοργανισμού.\\γ. γενετικά τροποποιημένα βακτήρια.\\δ. αδρανοποιημένους ιούς.
    \item[\textlatin{Meltemi:}] α.

    \item[\textlatin{User:}] Ποιος ήταν ο πρώτος που πραγματοποίησε πρακτικά, το 1803, την πρόωση των πλοίων με ατμοκίνηση;
    \item[\textlatin{Meltemi:}] Ο πρώτος που πραγματοποίησε πρακτικά την προώση των πλοίων με ατμοκίνηση ήταν ο Ρόμπερτ Φούλτον.
    
    \item[\textlatin{User:}] Από πόσους στίχους αποτελείται ο Ερωτόκριτος;
    \item[\textlatin{Meltemi:}] Ο Ερωτόκριτος αποτελείται από 10.012 στίχους.

    \item[\textlatin{User:}] Τι είναι τα συνελικτικά νευρωνικά δίκτυα;
    \item[\textlatin{Meltemi:}] Τα Συνελικτικά Νευρωνικά Δίκτυα (\textlatin{CNN}) είναι ένας τύπος αλγόριθμου μηχανικής μάθησης που είναι ιδιαίτερα κατάλληλος για εργασίες αναγνώρισης εικόνας και βίντεο. Είναι ένα υποσύνολο των νευρωνικών δικτύων που χρησιμοποιούν φίλτρα και στρώματα συγκέντρωσης για να εξάγουν χαρακτηριστικά από εικόνες ή ακολουθίες εισόδου βίντεο, καθιστώντας τα εξαιρετικά χρήσιμα στην ταξινόμηση εικόνων, τον εντοπισμό αντικειμένων, την ανίχνευση προσώπου, ακόμη και τη συμπίεση δεδομένων.
    
    \item[\textlatin{User:}] Ποιος ήταν ο Κατσουσίκα Χοκουσάι;
    \item[\textlatin{Meltemi:}] Ο Κατσουσίκα Χοκουσάι ήταν Ιάπωνας καλλιτέχνης που έζησε από το 1760 έως το 1849 και θεωρείται ένας από τους κορυφαίους καλλιτέχνες της εποχής Έντο στην Ιαπωνία. Είναι γνωστός για τις ξυλογραφίες του, συμπεριλαμβανομένης της διάσημης σειράς ``Τριάντα Έξι Όψεις του Φούτζι''. 
\end{description}

\end{document}